\documentclass[11pt,a4paper]{article}
\usepackage[hyperref]{acl2017}
\usepackage{amsmath}
\usepackage{amssymb}
\usepackage{bm}
\usepackage{bbm}
\usepackage{caption}
\usepackage{color}
\usepackage{graphicx}
\usepackage{lingmacros}

\usepackage{listings}
\usepackage{setspace}
\usepackage{comment}
\usepackage{times}
\usepackage{latexsym}
\usepackage{breqn}
\usepackage{url}

\usepackage{fancyvrb}
\VerbatimFootnotes

\newcommand{\norm}[1]{\left\lVert#1\right\rVert}

\aclfinalcopy 
\def\aclpaperid{777} 

\title{Other Topics You May Also Agree or Disagree:\\Modeling Inter-Topic Preferences using Tweets and Matrix Factorization}

\author{
  Akira Sasaki, Kazuaki Hanawa, Naoaki Okazaki, \and Kentaro Inui\\
  Graduate School of Information Sciences\\
  Tohoku University\\
  {\tt \{aki-s, hanawa, okazaki, inui\}@ecei.tohoku.ac.jp}
}

\date{}

\begin{document}
\maketitle
\begin{abstract}
  We present in this paper our approach for modeling inter-topic preferences of Twitter users: for example, {\it those who agree with the Trans-Pacific Partnership (TPP) also agree with free trade}.
  This kind of knowledge is useful not only for stance detection across multiple topics but also for various real-world applications including public opinion surveys, electoral predictions, electoral campaigns, and online debates.
  In order to extract users' preferences on Twitter, we design linguistic patterns in which people agree and disagree about specific topics (e.g., ``$\underline{A}$ is completely wrong'').
  By applying these linguistic patterns to a collection of tweets, we extract statements agreeing and disagreeing with various topics.
  Inspired by previous work on item recommendation, we formalize the task of modeling inter-topic preferences as matrix factorization: representing users' preferences as a user-topic matrix and mapping both users and topics onto a latent feature space that abstracts the preferences.
  Our experimental results demonstrate both that our proposed approach is useful in predicting missing preferences of users and that the latent vector representations of topics successfully encode inter-topic preferences.
\end{abstract}

\section{Introduction}
\label{sec:intro}

\begin{figure*}[t]
 \centering
 \includegraphics[width=1.0\hsize]{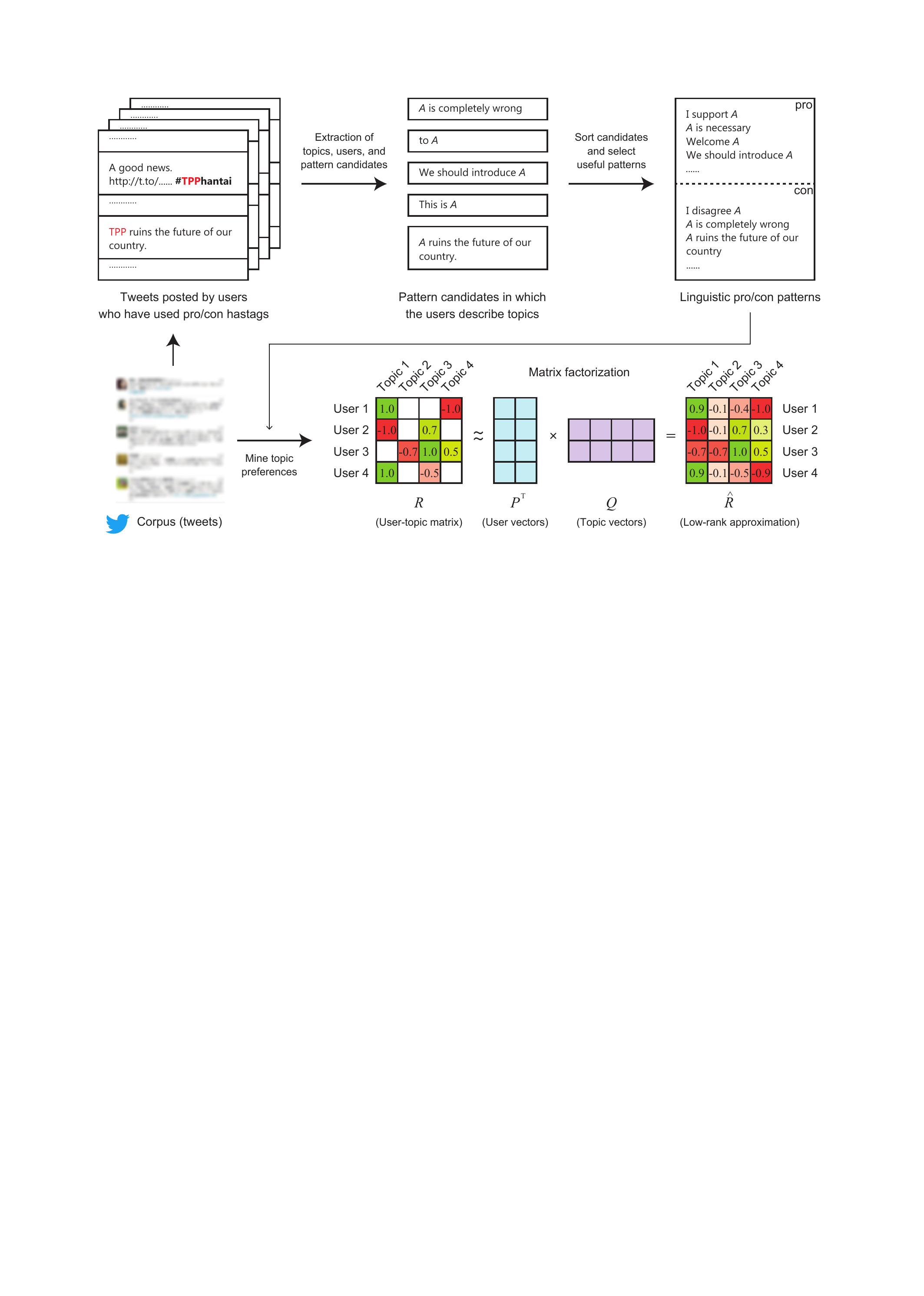}
 \caption{An overview of this study.}
 \label{fig:overall}
\end{figure*}

Social media have changed the way people shape public opinion.
The latest survey by the Pew Research Center reported that a majority of US adults (62\%) obtain news via social media, and of those, 18\% do so often~\cite{Gottfried:16}.
Given that news and opinions are shared and amplified by friend networks of individuals~\cite{Jamieson:08}, individuals are thereby isolated from information that does not fit well with their opinions~\cite{Pariser:11}.
Ironically, cutting-edge social media technologies promote ideological groups even with its potential to deliver diverse information.

A large number of studies already analyzed discussions, interactions, influences, and communities on social media along the political spectrum from liberal to conservative~\cite{Adamic:2005,zhou2011classifying,Cohen:13,Bakshy:15,Wong:2016}.
Even though these studies provide intuitive visualizations and interpretations along the liberal-conservative axis, political analysts argue that the axis is flawed and insufficient for representing public opinion and ideologies~\cite{Kerlinger:84,Maddox:84}.

A potential solution for analyzing multiple axes of the political spectrum on social media is stance detection~\cite{Thomas:2006,somasundaran2009recognizing,murakami2010support,Anand:2011,Walker:2012,Mohammad:2016,Johnson:2016}, whose task is to determine whether the author of a text is for, neutral, or against a topic (e.g., {\it free trade}, {\it immigration}, {\it abortion}).
However, stance detection across different topics is extremely difficult.
\newcite{Anand:2011} reported that a sophisticated method with topic-dependent features substantially improved the performance of stance detection within a topic, but such an approach could not outperform a baseline method with simple $n$-gram features when evaluated across topics.
More recently, all participants of SemEval 2016 Task 6A (with five topics) could not outperform the baseline supervised method using $n$-gram features~\cite{Mohammad:2016}.

In addition, stance detection encounters difficulties with different user types.
\newcite{Cohen:13} observed that existing methods on stance detection fail on ``ordinary'' users because such methods primarily obtain training and test data from politically vocal users (e.g., politicians);
for example, they found that a stance detector trained on a dataset with politicians achieved 91\% accuracy on other politicians but only achieved 54\% accuracy on ``ordinary'' users.
Establishing a bridge across different topics and users remains a major challenge not only in stance detection, but also in social media analytics.

An important component in establishing this bridge is commonsense knowledge about topics.
For example, consider a topic {\it a revision of Article 96 of the Japanese Constitution}.
We infer that the statement ``we should maintain armed forces'' tends to favor this topic even without any lexical overlap between the topic and the statement.
This inference is reasonable because: the writer of the statement favors {\it armed forces}; those who favor {\it armed forces} also favor {\it a revision of Article 9}\footnote{Article 9 prohibits armed forces in Japan.}; and those who favor {\it a revision of Article 9} also favor {\it a revision of Article 96}\footnote{Article 96 specifies high requirements for making amendments to Constitution of Japan (including Article 9).}.
In general, this kind of commonsense knowledge can be expressed in the format:
{\it those who agree/disagree with topic $A$ also agree/disagree with topic $B$}.
We call this kind of knowledge {\it inter-topic preference} throughout this paper.

We conjecture that previous work on stance detection indirectly learns inter-topic preferences within the same target through the use of $n$-gram features on a supervision data.
In contrast, in the present paper, we directly acquire inter-topic preferences from an unlabeled corpus of tweets.
This acquired knowledge regarding inter-topic preferences is useful not only for stance detection, but also for various real-world applications including public opinion survey, electoral campaigns, electoral predictions,  and online debates.

Figure \ref{fig:overall} provides an overview of this work.
In our system, we extract linguistic patterns in which people agree and disagree about specific topics (e.g., ``$\underline{A}$ is completely wrong''); to accomplish this, as described in Section \ref{sec:pattern-mining}, we make use of hashtags within a large collection of tweets.
The patterns are then used to extract instances of users' preferences regarding various topics, as detailed in Section \ref{sec:instance-extraction}.
Inspired by previous work on item recommendation, in Section \ref{sec:mf}, we formalize the task of modeling inter-topic preferences as a matrix factorization: representing a sparse user-topic matrix (i.e., the extracted instances) with the product of low-rank user and topic matrices.
These low-rank matrices provide {\it latent vector representations} of both users and topics.
This approach is also useful for completing preferences of ``ordinary'' (i.e., less vocal) users, which fills the gap between different types of users.

The contributions of this paper are threefold.
\begin{enumerate}
 \item To the best of our knowledge, this is the first study that models inter-topic preferences for unlimited targets on real-world data.
 \item Our experimental results show that this approach can accurately predict missing topic preferences of users accurately (80--94\%).
 \item Our experimental results also demonstrate that the latent vector representations of topics successfully encode inter-topic preferences, e.g., {\it those who agree with nuclear power plants also agree with nuclear fuel cycles}.
\end{enumerate}
This study uses a Japanese Twitter corpus because of its availability from the authors, but the core idea is applicable to any language.

\section{Mining Topic Preferences of Users}
\label{sec:pattern}

In this section, we describe how we collect statements in which users agree or disagree with various topics on Twitter, which then serves as source data for modeling inter-topic preferences.
More formally, we are interested in acquiring a collection of tuples $(u, t, v)$, where: $u \in U$ is a user; $U$ is the set of all users on Twitter; $t \in T$ is a topic; $T$ is the set of all topics; and $v \in \{+1,-1\}$ is $+1$ when the user $u$ agrees with the topic $t$ and $-1$ otherwise (i.e., disagreement).

Throughout this work, we use a corpus consisting of 35,328,745,115 Japanese tweets (7,340,730 users) crawled from February 6, 2013 to September 30, 2016.
We removed retweets from the corpus.

\subsection{Mining Linguistic Patterns of Agreement and Disagreement}
\label{sec:pattern-mining}

We use linguistic patterns to extract tuples $(u, t, v)$ from the aforementioned corpus.
More specifically, when a tweet message matches to one of linguistic patterns of agreement (e.g., ``$\underline{t}$ is necessary''), we regard that the author $u$ of the tweet agrees with topic $t$.
Conversely, a statement of disagreement is identified by linguistic patterns for disagreement (e.g., ``$\underline{t}$ is unacceptable'').

In order to design linguistic patterns, we focus on hashtags appearing in the corpus that have been popular clues for locating subjective statements such as sentiments~\cite{Davidov:2010}, emotions~\cite{Qadir:2014}, and ironies~\cite{VanHee:2016}.
Hashtags are also useful for finding strong supporters and critics, as well as their target topics; for example, \verb|#immigrantsWelcome| indicates that the author favors {\it immigrants}; and \verb|#StopAbortion| is against {\it abortion}.

Based on this intuition, we design regular expressions for both {\it pro hashtags} ``\verb|#(.+)sansei|''\footnote{Unlike English hashtags, we systematically attach a noun {\it sansei}, which stands for {\it pro} (agreement) in Japanese, to a topic, for example, \verb|#TPPsansei|. This paper uses the alphabetical expression \verb|sansei| only for explanation; the actual pattern uses Chinese characters corresponding to {\it sansei}.} and {\it con hashtags} ``\verb|#(.+)hantai|''\footnote{A Japanese noun {\it hantai} stands for {\it con} (disagreement), for example, \verb|#TPPhantai|. This paper uses the alphabetical expression \verb|hantai| only for explanation; the actual pattern uses Chinese characters corresponding to {\it hantai}.}, where \verb|(.+)| matches a target topic.
These regular expressions can find users who have strong preferences to topics.
Using this approach, we extracted 31,068 occurrences of pro/con hashtags used by 18,582 users for 4,899 topics.
We regard the set of topics found using this procedure as set of target topics $T$ in this study.

Each time we encounter a tweet containing a pro/con hashtag, we searched for corresponding textual statements as follows.
Suppose that a tweet includes a hashtag (e.g., \verb|#TPPsansei|) for a topic $t$ (e.g., {\it TPP}).
Assuming that the author of the given tweet does not change their attitude toward a topic over time, we search for other tweets posted by the same author that also have the topic keyword $t$.
This process retrieves tweets like ``I support TPP.''
Then, we replace the topic keyword into a variable $A$ to extract patterns, e.g., ``I support $\underline{A}$.''
Here, the definition of the pattern unit is language specific.
For Japanese tweets, we simply recognize a pattern that starts with a variable (i.e., topic) and ends at the end of the sentence\footnote{In English, this treatment roughly corresponds to extracting a verb phrase with the variable $A$.}.

Because this procedure also extracts useless patterns such as ``to $A$'' and ``this is $A$'', we manually choose useful patterns in a systematic way:
sort patterns in descending order of the number of users who use the pattern; and check the sorted list of patterns manually; and remove useless patterns.
Using this approach, we obtained 100 pro patterns (e.g., ``welcome $A$'' and ``$A$ is necessary'') and 100 con patterns (``do not let $A$'' and ``I don't want $A$'').

\subsection{Extracting Instances of Topic Preferences}
\label{sec:instance-extraction}

By using the pro and con patterns acquired using the approach described in Section \ref{sec:pattern-mining}, we extract instances of $(u, t, v)$ as follows.
When a sentence in a tweet whose author is user $u$ matches one of the pro patterns (e.g., ``{\it $\underline{t}$ is necessary}'') and the topic $t$ is included in the set of target topics $T$, we recognize this as an instance of $(u, t, +1)$.
Similarly, when a sentence matches one of the con patterns (e.g., ``I don't want $\underline{t}$'') and the topic $t$ is included in the set of target topics $T$, we recognize this as an instance of $(u, t, -1)$.
Using this approach, we collected 25,805,909 tuples corresponding to 3,302,613 users and 4,899 topics.
Because these collected tuples included comparatively infrequent users and topics, we removed users and topics that appeared less than five times.
In addition, there were also meaningless frequent topics such as ``of'' and ``it''.
Therefore, we sorted topics in descending order of their co-occurrence frequencies with each of the pro patterns and con patterns,
and then removed meaningless topics in the top 100 topics.
This resulted in 9,961,509 tuples regarding 273,417 users and 2,323 topics.

\section{Matrix Factorization}
\label{sec:mf}

Using the methods described in Section \ref{sec:pattern}, from the corpus, we collected a number of instances of users' preferences regarding various topics.
However, Twitter users do not necessarily express preferences for all topics.
In addition, it is by nature impossible to predict whether a new (i.e., nonexistent in the data) user agrees or disagrees with given topics.
Therefore, in this section, we apply matrix factorization~\cite{Koren:2009} in order to predict missing values, inspired by research regarding item recommendation~\cite{Bell:2007,Dror:2011}.
In essence, matrix factorization maps both users and topics onto a latent feature space that abstracts topic preferences of users.

Here, let $R$ be a sparse matrix of $|U| \times |T|$.
Only when a user $u$ expresses a preference for topic $t$ do we compute an element of the sparse matrix $r_{u,t}$,
\begin{gather}
 r_{u,t} = \frac{\#(u, t, +1) - \#(u, t, -1)}{\#(u, t, +1) + \#(u, t, -1)} \label{equ:user-target-matrix}
\end{gather}
Here, $\#(u, t, +1)$ and $\#(u, t, -1)$ represent the numbers of occurrences of instances $(u, t, +1)$ and $(u, t, -1)$, respectively.
Thus, an element $r_{u, t}$ approaches $+1$ as the user $u$ favors the topic $t$, and $-1$ otherwise.
If the user $u$ does not make any statement regarding the topic $t$ (i.e., neither $(u, t, +1)$ nor $(u, t, -1)$ exists in the data), we do not fill the corresponding element, leaving it as a missing value.

Matrix factorization decomposes the sparse matrix $R$ into low-dimensional matrices $P \in \mathbb{R}^{k \times |U|}$ and $Q \in \mathbb{R}^{k \times |T|}$, where $k$ is a parameter that specifies the number of dimensions of the latent space.
We minimize the following objective function to find the matrices $P$ and $Q$,
\begin{align}
  \min_{P, Q} \sum_{(u,t) \in R} \biggl( (r_{u,t} - {{\bm p}_u}^{\intercal} {\bm q}_t)^2 \nonumber \\ + \lambda_{P}\norm{{\bm p}_u}^2 + \lambda_{Q}\norm{{\bm q}_t}^2 \biggr) .
  \label{eq:objective}
\end{align}
Here, $(u, t) \in R$ is repeated for elements filled in the sparse matrix $R$,
${\bm p}_u \in \mathbb{R}^{k}$ and ${\bm q}_v \in \mathbb{R}^{k}$ are $u$ column vectors of $P$ and $v$ column vectors of $Q$, respectively,
and $\lambda_{P} \geq 0$ and $\lambda_{Q} \geq 0$ represent coefficients of regularization terms.
We call ${\bm p}_u$ and ${\bm q}_t$ the {\it user vector} and {\it topic vector}, respectively.

Using these user and topic vectors, we can predict an element $\hat{r}_{u,t}$ that may be missing in the original matrix $R$,
\begin{gather}
    \hat{r}_{u,t} \simeq {{\bm p}_u}^{\intercal} {\bm q}_t \label{equ:prediction}.
\end{gather}
We use {\tt libmf}\footnote{\url{https://github.com/cjlin1/libmf}}~\cite{chin2015fast} to solve the optimization problem in Equation \ref{eq:objective}.
We set regularization coefficients $\lambda_{P} = 0.1$ and $\lambda_{Q} = 0.1$ and use default values for the other parameters of {\tt libmf}.

\section{Evaluation}
\label{sec:ex}

\subsection{Determining the Dimension Parameter $k$}

\begin{figure}[t]
 \centering
 \includegraphics[width=0.81\hsize]{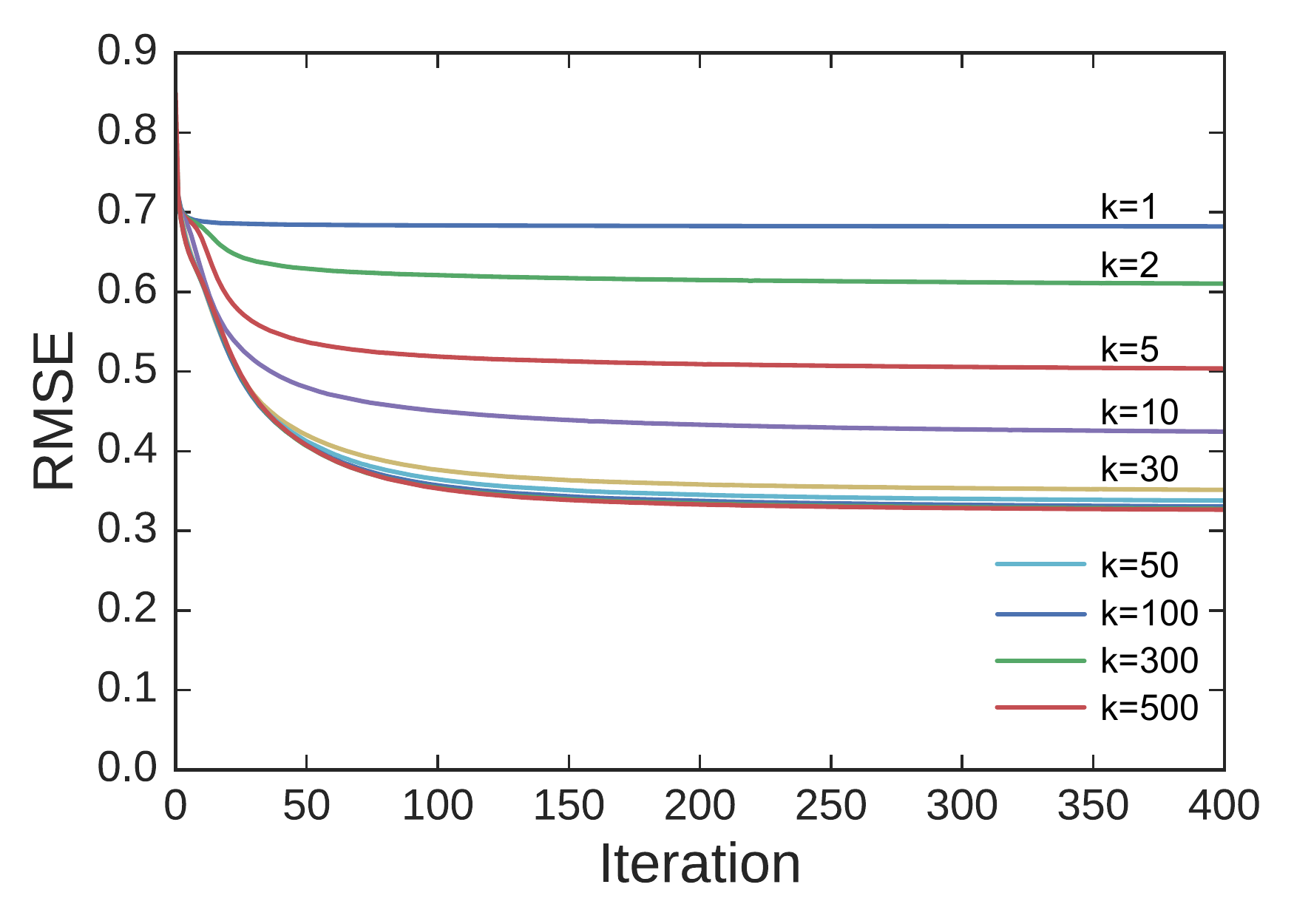}
 \caption{Reconstruction error (RMSE) of matrix factorization with different $k$.}
 \label{fig:rmse}
\end{figure}

How good is the low-rank approximation found by matrix factorization?
And can we find the ``sweet spot'' for the number of dimensions $k$ of the latent space?
We investigate the reconstruction error of matrix factorization using different values of $k$ to answer these questions.
We use Root Mean Squared Error (RMSE) to measure error,
\begin{gather}
  RMSE=\sqrt{\frac{\sum_{(u,t) \in R}{({{\bm p}_u}^{\intercal} {\bm q}_t - r_{u,t})^2}}{N}} .
\end{gather}
Here, $N$ is the number of elements in the sparse matrix $R$ (i.e., the number of known values).

Figure \ref{fig:rmse} shows RMSE values over iterations of {\tt libmf} with the dimension parameter $k \in \{1, 2, 5, 10, 30, 50, 100, 300, 500\}$.
We observed that the reconstruction error decreased as the iterative method of {\tt libmf} progressed.
The larger the number of dimensions $k$ was, the smaller the reconstruction error became; the lowest reconstruction error was 0.3256 with $k=500$.
We also observed the error with $k=1$, which corresponds to mapping users and topics onto one dimension similarly to the political spectrum of liberal and conservative.
Judging from the relatively high RMSE values with $k=1$, we conclude that it may be difficult to represent everything in the data using a one-dimensional axis.
Based on this result, we concluded that matrix factorization with $k=100$ is sufficient for reconstructing the original matrix $R$ and therefore used this parameter value for the rest of our experiments.

\subsection{Predicting Missing Topic Preferences}

How accurately can the user and topic vectors predict missing topic preferences?
To answer this question, we evaluate the accuracy in predicting hidden preferences in the matrix $R$ as follows.
\begin{figure}[ht]
 \centering
 \includegraphics[width=1.05\hsize]{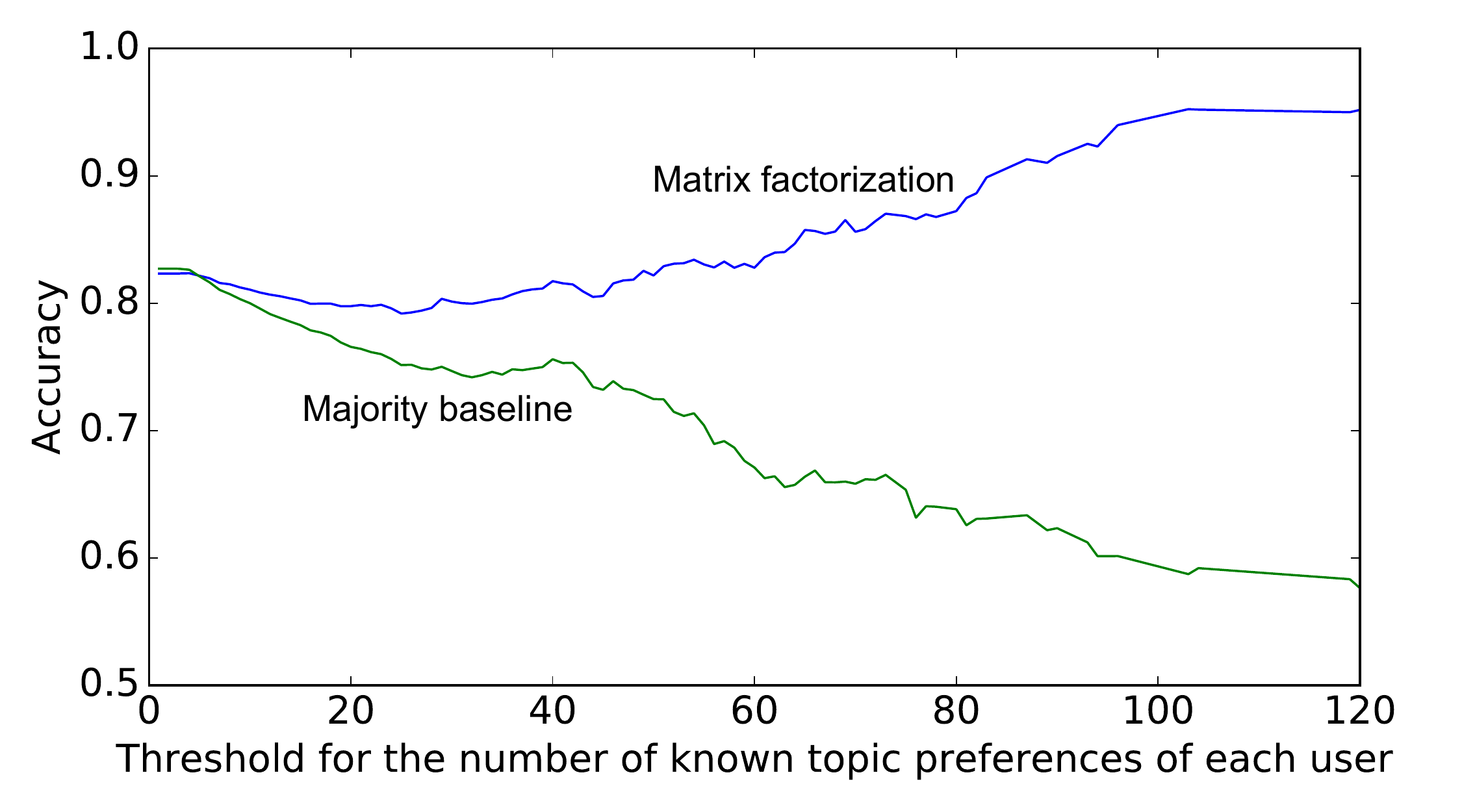}
 \caption{Prediction accuracy when changing the threshold for the number of known topic preferences of each user.}
 \label{fig:predict_5}
\end{figure}
First, we randomly selected 5\% of existing elements in $R$ and let $Y$ represent the collection of the selected elements (test set).
We then perform matrix factorization on the sparse matrix without the selected elements of $Y$, that is, only with the remaining 95\% elements of $R$ (training set).
We define the accuracy of the prediction as
\begin{align}
\frac{1}{|Y|} \sum_{u, t \in Y} \mathbbm{1} \left({\rm sign}(\hat{r}_{u,t}) = {\rm sign}(r_{u,t})\right) \label{equ:accuracy}
\end{align}
Here, $r_{u,t}$ denotes the actual (i.e., self-declared) preference values, $\hat{r}_{u,t}$ represents the preference value predicted by Equation \ref{equ:prediction}, ${\rm sign}(.)$ represents the sign of the argument, and $\mathbbm{1}(.)$ yields $1$ only when the condition described in the argument holds and $0$ otherwise.
In other words, Equation \ref{equ:accuracy} computes the proportion of correct predictions to all predictions, assuming zero to be the decision boundary between pro and con.

Figure \ref{fig:predict_5} plots prediction accuracy values calculated from different sets of users.
Here the x-axis represents a threshold $\theta$, which filters out users whose declarations of topic preferences are no greater than $\theta$ topics.
In other words, Figure \ref{fig:predict_5} shows prediction accuracy when we know user preferences for at least $\theta$ topics.
For comparison, we also include the majority baseline that predicts pro and con based on the majority of preferences regarding each topic in the training set.

Our proposed method was able to predict missing preferences with an 82.1\% accuracy for users stating preferences for at least five topics.
This accuracy increased as our method received more information regarding the users, reaching a 94.0\% accuracy when $\theta = 100$.
This result again indicates that our proposed method reasonably utilizes known preferences to complete missing preferences.

\begin{figure}[t]
 \centering
 \includegraphics[width=1.0\hsize]{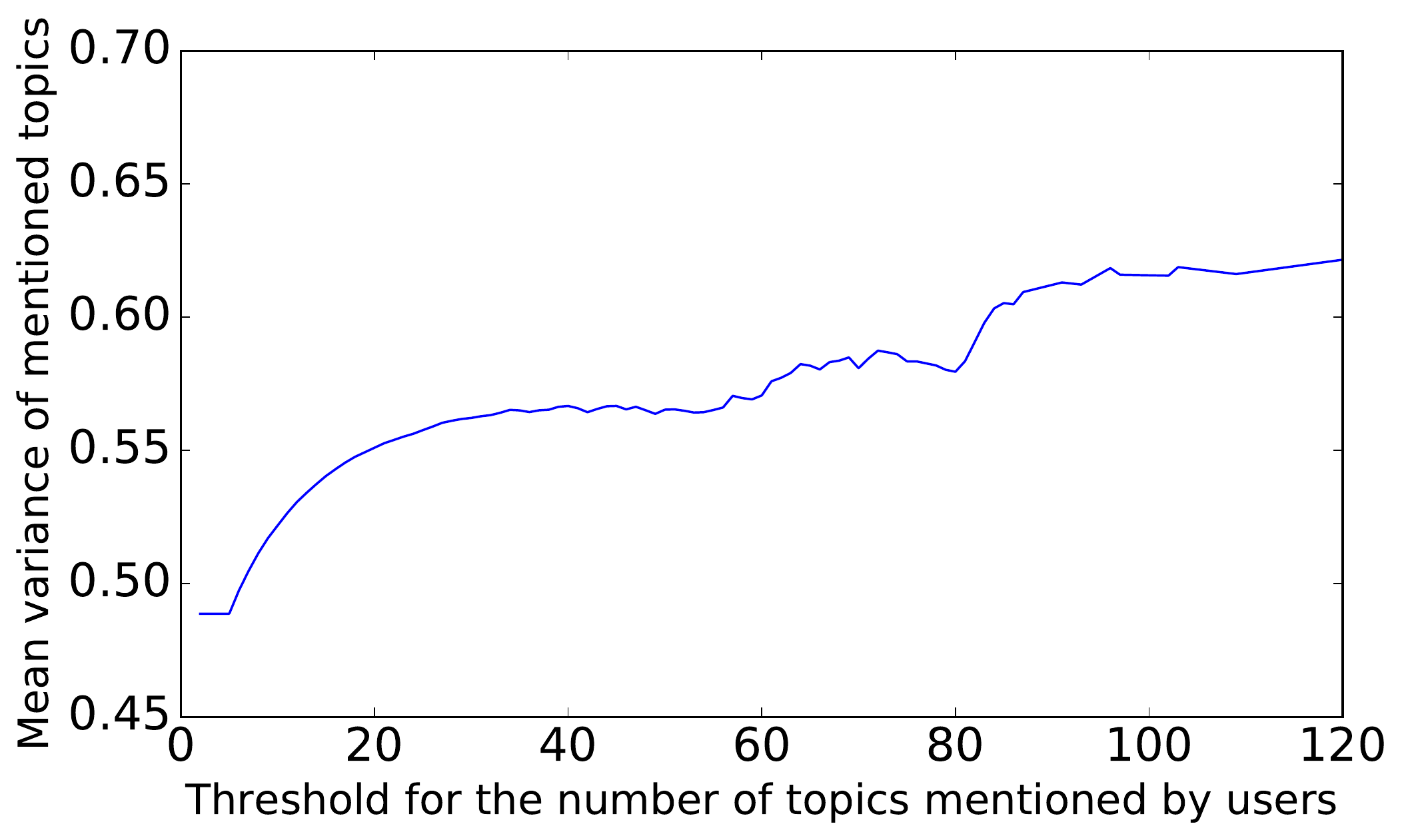}
 \caption{Mean variance of preference values of self-declared topics when changing the threshold for the number of self-declared topics.}
 \label{fig:variance}
\end{figure}

In contrast, the performance of the majority baseline decreased as it received more information regarding the users.
Because this result was rather counter-intuitive, we examined the cause of this phenomenon.
Consequently, this result turned out to be reasonable because preferences of vocal users deviated from those of the average users.
Figure \ref{fig:variance} illustrates this finding, showing the mean of variances of preference values $r_{u,t}$ across self-declared topics.
In the figure, the x-axis represents a threshold $\theta$, which filters out users whose statements of topic preferences are no greater than $\theta$ topics.
We observe that the mean variance increased as we focused on vocal users.
Overall, these results demonstrate the usefulness of user and topic vectors in predicting missing preferences.

\begin{table*}[t]
    \begin{center}
        \footnotesize
        \begin{tabular}{|l|l|p{10.8cm}|} \hline
            \multicolumn{1}{|c|}{User} & \multicolumn{1}{c|}{Type} & \multicolumn{1}{c|}{Topic} \\ \hline \hline
              A & Agreement (declared) & regime change, capital relocation \\ \cline{2-3}
                & Disagreement (declared) & Okinawa US military base, nuclear weapons, TPP, Abe Cabinet, Abe government, nuclear cycle, right to collective defense, nuclear power plant, Abenomics \\ \cline{2-3}
                & Agreement (predicted) & same-sex partnership ordinance (0.9697), vote of non-confidence to Cabinet (0.9248), national people's government (0.9157), abolition of tax (0.8978) \\ \cline{2-3}
                & Disagreement (predicted) & steamrollering war bill (-1.0522), worsening dispatch law (-1.0301), Sendai nuclear power plant (-1.0269), war bill (-1.0190), construction of a new base (-1.0186), Abe administration (-1.0173), landfill Henoko (-1.0158), unreasonable arrest (-1.0113) \\ \hline
              B & Agreement (declared) & visit shrine, marriage \\ \cline{2-3}
                & Disagreement(declared) & tax increase, conscription, amend Article 9 \\ \cline{2-3}
                & Agreement (predicted) & national people's government (0.8467), abolition of tax (0.8300), same-sex partnership ordinance (0.7700), security bills (0.6736) \\ \cline{2-3}
                & Disagreement (predicted) & corporate tax cuts (-1.0439), Liberal Democratic Party's draft constitution (-1.0396), radioactivity (-1.0276), rubble (-1.0159), nuclear cycle (-1.0143) \\ \hline
        \end{tabular}
        \caption{Examples of agreement/disagreement topics predicted for two sample users A and B, with predicted score $\hat{r}_{u, v}$ shown in parenthesis.}
        \label{tb:qualitative}
    \end{center}
\end{table*}

Table \ref{tb:qualitative} shows examples in which missing preferences of two users were predicted from known statements of agreements and disagreements\footnote{We anonymized user names in these examples. In addition, we removed topics that are too discriminatory or aggressive to other countries and races. Even though the experimental results of this paper do not necessarily reflect our idea, we do not think it is a good idea to distribute politically incorrect ideas through this paper.}.
In the table, predicted topics are accompanied by the corresponding $\hat{r}_{u,t}$ value in parentheses.
As an example, our proposed method predicted that the user A, who is positive toward {\it regime change} but negative toward {\it Okinawa US military base}, may also be positive toward {\it vote of non-confidence to Cabinet} but negative toward {\it construction of a new base}.

\subsection{Inter-topic Preferences}

\begin{table*}[t]
    \begin{center}
        \footnotesize
        \begin{tabular}{|l|p{11.0cm}|} \hline
            \multicolumn{1}{|c|}{Topic} & \multicolumn{1}{c|}{Topics with a high degree of cosine similarity} \\ \hline \hline
            Liberal Democratic Party (LDP) & Abe's LDP (0.3937), resuming nuclear power plant operations (0.3765), bus rapid transit (BRT) (0.3410), hate speech countermeasure law (0.3373), Henoko relocation (0.3353), C-130 (0.3338), Abe administration (0.3248), LDP \& Komeito (0.2898), Prime Minister Abe (0.2835) \\ \hline
            constitutional amendment & amendment of Article 9 (0.4520), enforcement of specific secret protection law (0.4399), security related law (0.4242), specific confidentiality protection law (0.4022), security bill amendment (0.3977), defense forces (0.3962), my number law (0.3874), collective self-defense rights (0.3687), militarist revival (0.3567) \\ \hline
            right of foreigners to vote & human rights law (0.5405), anti-discrimination law (0.5376), hate speech countermeasure law (0.5080), foreigner's life protection (0.4553), immigration refugee (0.4520), co-organized Olympics (0.4379) \\ \hline
        \end{tabular}
        \caption{Topics identified as being similar to the three controversial topics shown in the left column.}
        \label{tb:cossim}
    \end{center}
\end{table*}

Do the topic vectors obtained by matrix factorization
capture inter-topic preferences, such as ``People who agree with A also agree with B''?

Because no dataset exists for this evaluation, we created a dataset of pairwise inter-topic preferences by using a crowdsourcing service\footnote{We used Yahoo! Crowdsourcing, a Japanese online service for crowdsourcing.\\\url{http://crowdsourcing.yahoo.co.jp/}}.
Sampling topic pairs randomly, we collected 150 topic pairs whose cosine similarities of topic vectors were below $-0.6$, 150 pairs whose cosine similarities were between $-0.6$ and $0.6$, and 150 pairs whose cosine similarities were above $0.6$.
In this way, we obtained 450 topic pairs for evaluation.

Given a pair of topics $A$ and $B$, a crowd worker was asked to choose a label from the following three options:
(a) {\it those who agree/disagree with topic $A$ may also agree/disagree with topic $B$};
(b) {\it those who agree/disagree with topic $A$ may conversely disagree/agree with topic $B$};
(c) {\it otherwise (no association between $A$ and $B$)}.
Creating twenty pairs of topics as gold data, we removed labeling results from workers whose accuracy is less than 90\%.

Consequently, we obtained 6--10 human judgements for every topic pair.
Regarding (a) as $+1$ point, (b) as $-1$ point, and (c) as $0$ point, we computed the mean of the points (i.e., average human judgements) for each topic pair.
Spearman's rank correlation coefficient ($\rho$) between cosine similarity values of topic vectors and human judgements was $0.2210$.
We could observe a moderate correlation even though inter-topic preferences collected in this manner were highly subjective.

In addition to the quantitative evaluation, as summarized in Table \ref{tb:cossim}, we also checked similar topics for three controversial topics, {\it Liberal Democratic Party (LDP)}, {\it constitutional amendment} and {\it right of foreigners to vote} (Table \ref{tb:cossim}).
Topics similar to LDP included synonymous ones (e.g., {\it Abe's LDP} and {\it Abe administration}) and
other topics promoted by the {\it LDP} (e.g., {\it resuming nuclear power plant operations}, {\it bus rapid transit (BRT)} and {\it hate speech countermeasure law}).
Considering that people who support the LDP may also tend to favor its policies, we found these results reasonable.
As for the other example, {\it constitutional amendment} had a feature vector that was similar to that of {\it amendment of Article 9}, {\it enforcement of specific secret protection law} and {\it security related law}.
From these results, we concluded that topic vectors were able to capture inter-topic preferences.

\section{Related Work}

In this section, we summarize the related work that spreads across various research fields.

\paragraph{Social Science and Political Science}

A number of of studies analyze social phenomena regarding political activities, political thoughts, and public opinions on social media.
These studies model the political spectrum from liberal to conservative~\cite{Adamic:2005,zhou2011classifying,Cohen:13,Bakshy:15,Wong:2016}, political parties~\cite{tumasjan2010predicting,boutet2013s,Makazhanov:2013}, and elections~\cite{OConnor:2010,conover2011predicting}.

Employing a single axis (e.g., liberal to conservative) or a few axes (e.g., political parties and candidates of elections), these studies provide intuitive visualizations and interpretations along the respective axes.
In contrast, this study is the first attempt to recognize and organize various axes of topics on social media with no prior assumptions regarding the axes.
Therefore, we think our study provides a new tool for computational social science and political science that enables researchers to analyze and interpret phenomena on social media.

Next, we describe previous research focused on acquiring lexical knowledge of politics.
\newcite{Sim:2013} measured ideological positions of candidates in US presidential elections from their speeches.
The study first constructs ``cue lexicons'' from political writings labeled with ideologies by domain experts, using sparse additive generative models~\cite{Eisenstein:2011}.
These constructed cue lexicons were associated with such ideologies as {\it left}, {\it center}, and {\it right}.
Representing each speech of a candidate with cue lexicons, they inferred the proportions of ideologies of the candidate.
The study requires a predefined set of labels and text data associated with the labels.

\newcite{Bamman:2015} presented an unsupervised method for assessing the political stance of a proposition, such as ``global warming is a hoax,'' along the political spectrum of liberal to conservative.
In their work, a proposition was represented by a tuple in the form $\langle \mbox{subject}, \mbox{predicate}\rangle$, for example, $\langle \mbox{\it global warming}, \mbox{\it hoax}\rangle$.
They presented a generative model for users, subjects, and predicates to find a one-dimensional latent space that corresponded to the political spectrum.

Similar to our present work, their work~\cite{Bamman:2015} did not require labeled data to map users and topics (i.e., subjects) onto a latent feature space.
In their paper, they reported that the generative model outperformed Principal Component Analysis (PCA), which is a method for matrix factorization.
Empirical results here probably reflected the underlying assumptions that PCA treats missing elements as zero and not as missing data.
In contrast, in the present work, we properly distinguish missing values from zero, excluding missing elements of the original matrix from the objective function of Equation \ref{eq:objective}.
Further, this work demonstrated the usefulness of the latent space, that is, topic and user vectors, in predicting missing topic preferences of users and inter-topic preferences.

\paragraph{Fine-grained Opinion Analysis}

The method presented in Section \ref{sec:pattern} is an instance of fine-grained opinion analysis~\cite{Wiebe2005,Choi:2006,Johansson:2010,Yang:2013,Deng:2015}, which extracts a tuple of a subjective opinion, a holder of the opinion, and a target of the opinion from text.
Although these previous studies have the potential to improve the quality of the user-topic matrix $R$, unfortunately, no corpus or resource is available for the Japanese language.
We do not currently have a large collection of English tweets, but combining fine-grained opinion analysis with matrix factorization is an immediate future work.

\paragraph{Causality Relation}

Some of inter-topic preferences in this work can be explained by causality relation, for example, ``TPP promotes free trade.''
A number of previous studies acquire instances of causal relation~\cite{Girju:2003,Do:2011} and promote/suppress relation~\cite{hashimoto2012excitatory,Fluck:15} from text.
The causality knowledge is useful for predicting (hypotheses of) future events~\cite{Radinsky:2012,Radinsky:2012a,Hashimoto:2015}.

Inter-topic preferences, however, also include pairs of topics in which causality relation hardly holds.
As an example, it is unreasonable to infer that {\it nuclear plant} and {\it railroading of bills} have a causal relation, but those who dislike {\it nuclear plant} also oppose {\it railroading of bills} because presumably they think the governing political parties rush the bill for resuming a nuclear plant.
In this study, we model these inter-topic preferences based on preferences of the public.
That said, we have as a promising future direction of our work plans to incorporate approaches to acquire causality knowledge.

\section{Conclusion}
\label{sec:conclusion}

In this paper, we presented a novel approach for modeling inter-topic preferences of users on Twitter.
Designing linguistic patterns for identifying support and opposition statements, we extracted users' preferences regarding various topics from a large collection of tweets.
We formalized the task of modeling inter-topic preferences as a matrix factorization that maps both users and topics onto a latent feature space that abstracts users' preferences.
Through our experimental results, we demonstrated that our approach was able to accurately predict missing topic preferences of users (80--94\%) and that our latent vector representations of topics properly encoded inter-topic preferences.

For our immediate future work, we plan to embed the topic and user vectors to create a cross-topic stance detector.
It is possible to generalize our work to model heterogeneous signals, such as interests and behaviors of people, for example, ``those who are interested in $A$ also support $B$,'' and ``those who favor $A$ also vote for $B$''.
Therefore, we believe that our work will bring about new applications in the field of NLP and other disciplines.

\section*{Acknowledgements}

This work was supported by JSPS KAKENHI Grant Number 15H05318 and JST CREST Grant Number J130002054, Japan.

\bibliography{acl2017}
\bibliographystyle{acl_natbib}

\appendix

\end{document}